\documentclass[conference]{IEEEtran}
\IEEEoverridecommandlockouts

\usepackage{amsmath, amssymb,amsfonts}
\usepackage{algorithmic}
\usepackage{graphicx}
\usepackage{textcomp}
\def\BibTeX{{\rm B\kern-.05em{\sc i\kern-.025em b}\kern-.08em
    T\kern-.1667em\lower.7ex\hbox{E}\kern-.125emX}}

\usepackage{graphicx}  
\usepackage[backend=bibtex, sorting=none, bibstyle=ieee]{biblatex}  
\usepackage{rotating}  
\usepackage[normalem]{ulem}  

\usepackage{amsmath}
\usepackage{amssymb}
\usepackage{amsbsy}
\usepackage{latexsym}
\usepackage{float}
\usepackage{epsfig}
\usepackage{epstopdf}
\usepackage{caption}
\usepackage{subcaption}
\usepackage{graphicx}
\usepackage{longtable}
\usepackage{tabu}
\usepackage{tabularx}
\usepackage{array}
\usepackage{grffile}
\usepackage{amsbsy}
\usepackage{rotating}
\usepackage{multirow}
\usepackage{multicol}
\usepackage{pdflscape}

\usepackage{xifthen}
\usepackage{xspace}
\usepackage{lipsum}
\usepackage{upgreek}
\usepackage{color}
\usepackage{url}
\usepackage{booktabs}

\usepackage{siunitx}
\usepackage[usenames,dvipsnames,svgnames,table]{xcolor} 
\usepackage{amsthm}
\usepackage{blindtext}

\bibliography{B_bib}

\AtEveryBibitem{\clearfield{issn}}
\AtEveryBibitem{\clearlist{issn}}

\AtEveryBibitem{\clearfield{language}}
\AtEveryBibitem{\clearlist{language}}

\AtEveryBibitem{\clearfield{doi}}
\AtEveryBibitem{\clearlist{doi}}

\AtEveryBibitem{\clearfield{url}}
\AtEveryBibitem{\clearlist{url}}

\AtEveryBibitem{%
  \ifentrytype{online}
    {}
    {\clearfield{urlyear}\clearfield{urlmonth}\clearfield{urlday}}}

\begin{document}



\title{Locomoting robots composed of immobile robots
\thanks{This research was funded by the U.S. Army Research Office.}}

\author{\IEEEauthorblockN{Ross Warkentin}
\IEEEauthorblockA{\textit{Woodruff School of Mechanical Engineering} \\
\textit{Georgia Institute of Technology}\\
Atlanta, U.S.A. \\
rwarkentin3@gatech.edu}
\and
\IEEEauthorblockN{William Savoie}
\IEEEauthorblockA{\textit{School of Physics} \\
\textit{Georgia Institute of Technology}\\
Atlanta, U.S.A. \\
wsavoie@gatech.edu}
\and
\IEEEauthorblockN{Daniel I. Goldman}
\IEEEauthorblockA{\textit{School of Physics} \\
\textit{Georgia Institute of Technology}\\
Atlanta, U.S.A. \\
daniel.goldman@physics.gatech.edu}
}
\maketitle

\begin{abstract}
Robotic materials are multi-robot systems formulated to leverage the low-order computation and actuation of the constituents to manipulate the high-order behavior of the entire material. We study the behaviors of ensembles composed of smart active particles, \textit{smarticles}. Smarticles are small, low cost robots equipped with basic actuation and sensing abilities that are individually incapable of rotating or displacing. We demonstrate that a ``supersmarticle'', composed of many smarticles constrained within a bounding membrane, can harness the  internal collisions of the robotic material among the constituents and the membrane to achieve diffusive locomotion. The emergent diffusion can be directed by modulating the robotic material properties in response to a light source, analogous to biological phototaxis. The light source introduces asymmetries within the robotic material, resulting in modified populations of interaction modes and dynamics which ultimately result in supersmarticle biased locomotion. We present experimental methods and results for the robotic material which moves with a directed displacement in response to a light source.
\end{abstract}


The task of robotic material design is both arduous and lucrative due to the tight integration of sensing, actuation, and computation capabilities which allow the material to utilize external stimuli to change the physical properties in a programmable fashion \cite{McEvoy2015}. Various forms of robot materials have been explored \cite{McEvoy2015}, such as the continuous materials which emulate biological systems like the camouflage patterns seen in cuttlefish and chameleons \cite{Stevens2009} or the dynamic shape-changing wings of birds \cite{Thill2008}. The integration of sensors, controllers, and actuators inspired by biological systems yields multifunctional materials with responsive properties that have immense potential in future technologies.

Examples of more abstract, discrete material ensembles can be found in the field of modular robotics. Modular self-reconfigurable robotic systems exploit motion planning and control of kinematic robotic subsystems to produce materials with controllable dynamic morphology \cite{Yim2007, Yu2007}. Coordination of the individual robots within the ensemble is often supported by powerful computational and communication capabilities that permit the material to operate cohesively.

Attention has been given to off-loading computation from the controller of the robot/ensemble to the mechanical components of the system (i.e., moving computation from the brain to the body). Morphological control systems modulate the system state space topology and moderate the movement from one attractor basin to another \cite{Fuchslin2013}. The morphology of the ensemble can also be used to communicate information within the robotic material, akin to the environment-based interactions (i.e., stigmergy) utilized by decentralized swarm robotic systems \cite{Beckers2000, Steels1990}. Interparticle regions within a robotic material provide a medium through which the agents of the bulk can interact.

To discover the principles of morphological computation with regards to motion, we study a robotic material of individually immobile particles which achieve a directed locomotion with respect to a light source, a phenomenon known as phototaxing. A control system is proposed which relies on the robotic material morphology to favor locomoting interaction modes and communicate neighboring states within the material by light occlusion. We treat the supersmarticle as a model for a robotic material which produces locomotion from dynamics which are entirely absent for individual agents, with attention given to controlling the material by modulating the individual agent properties.

\section{Robotic Material Implementation}
Robotic materials couple the morphology and computational capabilities of the system to autonomously respond to stimuli. These materials are defined by the presence of four principle constituent parts: (1) sensors to identify and respond to environmental changes, (2) actuators to change the properties of the material, (3) communication elements to transport sensory and control signals, and (4) computing elements to receive sensor signals and send actuator and communication signals \cite{McEvoy2015}. We first explore the smarticle design and the robotic material sensors, actuators, and internal computation are realized within the individual, before moving on to the supersmarticle where the elements of communication and self-organization are realized.

We propose a robotic material composed of collections of smart particles, \textit{smarticles}, with limited mobility and sensing capabilities.  Each smarticle is a small, 3-link, planar robot equipped with sensing abilities which is incapable of rotating or displacing individually. Each smarticle measures 14 x 2.5 x 3 $cm^3$, where only the center link is in contact with the ground. The links of the smarticles were 3D printed to ensure uniform construction between all smarticles. A supersmarticle is a collection of smarticles enclosed by an unanchored rigid ring. A supersmarticle is a robot material made of robots, leveraging the simple sensory and actuation capabilities of the individual smarticle in conjunction with the interaction dynamics of the ensemble to realize dynamics impossible for a single smarticle.

\begin{figure}[htbp]
\centerline{\includegraphics[width=.4\textwidth]{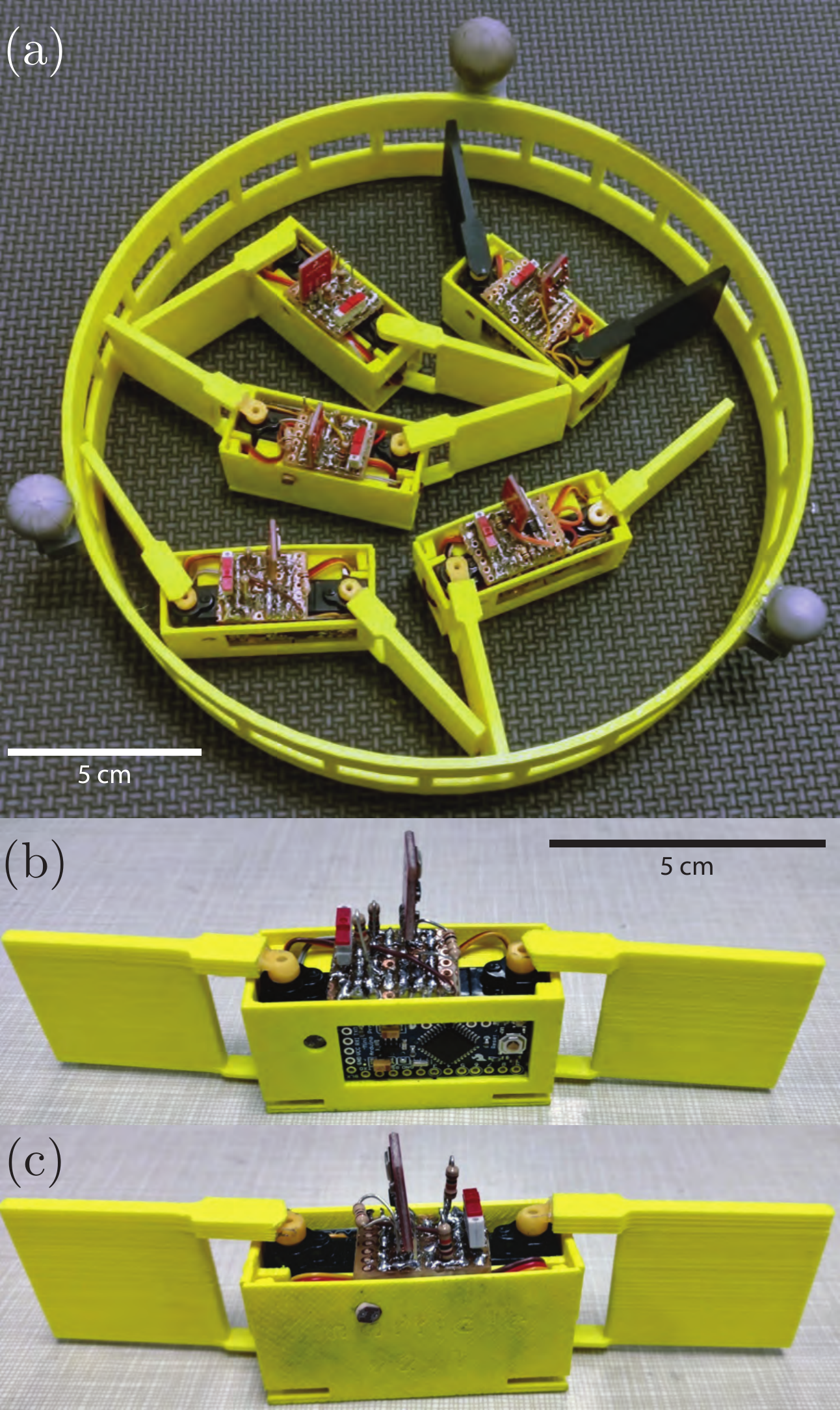}}
\caption{Supersmarticle is composed of many, confined, smarticles. (a) A supersmarticle composed of 5 individual
smarticles. A single smarticle, as viewed from the (b)
front and (c) rear.}
\label{fig:smarticles}
\end{figure}

Each smarticle has a MEMS analog omnidirectional microphone, two photoresistors, and a current sensing resistor in series with the servos which allows the smarticle to sense external and internal states. The microphone and photoresistors provide the smarticle the ability to respond to two unique, environmental stimuli we can programmatically modulate (tone frequency or light exposure) to send basic commands to the material. The current sensing resistor detects the current the servos draw, which is proportional to the torque the servos experience, allowing each smarticle to sense the local stress state.

\begin{figure}[htbp]
\centerline{\includegraphics[width=.5\textwidth]{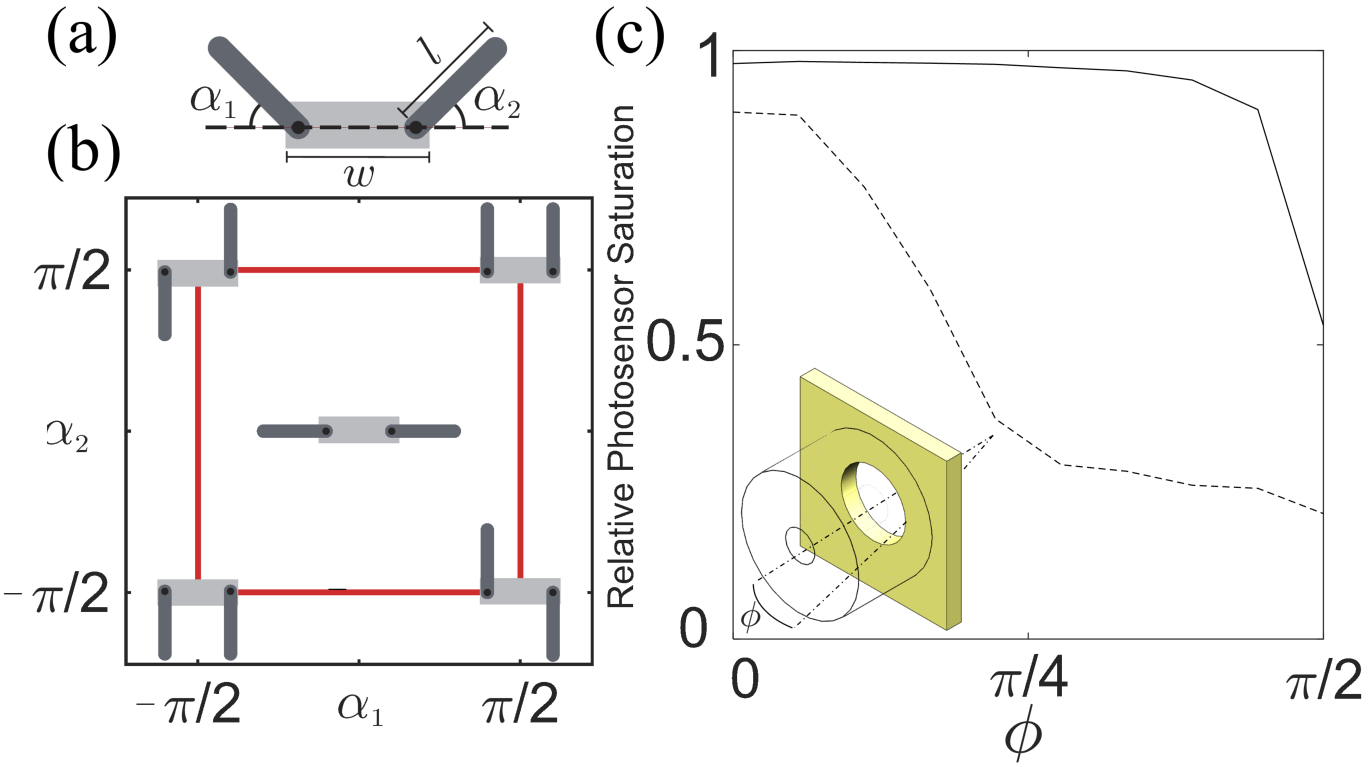}}
\caption{(a) Configuration space of a single smarticle (top view) defined by the angles $\alpha_1$ and $\alpha_2$ between the outer and inner links. (b) The square gait with certain configurations from the trajectory illustrated. (c) Photosensor saturation levels for exposed photosensors (solid) and shrouded photosensors (dashed). Inset is a close-up isometric view of the photosensor hole and the outline of the additional geometric element which enshrouds the photosensor.}
\label{fig:configurationSpace}
\end{figure}

The two Power HD-1440A MicroServos actuate the smarticle links allowing the smarticle to achieve configurations in the two-dimensional joint space, as seen in Fig. \ref{fig:configurationSpace}(a,b). The smarticle can independently control the body geometry to modulate the degree of convexity. The body can align all links to produce a straight configuration, or ``curl'' into a u-shaped particle.

The signals from the sensors are detected by a re-programmable Arduino Pro Mini 328-3.3V/8MHz microcontroller, which performs computations that handle the ADC and servo control. A collection of smarticles is placed inside an unanchored ring; as the smarticles move through their periodic trajectories, collisions between neighboring smarticles and the ring generate net displacement of the ring.

Communication through the environment is possible through a process known as stigmergy, a social communication method for indirect coordination: changes in the environment by one agent elicits a response in other agents \cite{Theraulaz1999}. The supersmarticle is planar, such that a light source placed on one side of the platform will be occluded by the nearest smarticle, creating an internal light gradient. Given the light sensor locations on the smarticle body and the geometry of the straight configuration, typically only one smarticle at a time will detect the light, occluding the light and keeping the others smarticle photoresistors from light exposure. The internal light gradient across the supersmarticle  provides a decentralized, stigmergic communication method. Light sensors provide the smarticle a response mechanism to both external light and occluding neighbors. Communication within the robotic material is not explicitly performed using any form of hardware or wireless communication protocol; the environment of the ensemble is the communication pathway.

\section{Methods}

Supersmarticle tests were performed using two distinct smarticle bodies, the only difference being the smarticle body and the controlling light source. The first smarticle body has the photoresistors completely exposed, such that the photosensor readings would saturate independent from the angle of incidence. The exposed smarticle ensembles were controlled using a manually directed light source located at the center of the edge of the test plate. The effect of the shroud on photoresistor sensitivity can be seen in Fig. \ref{fig:configurationSpace}(c). The light source was directed towards the nearest exposed photoresistor, thereby rendering a single smarticle within the system inactive. The second smarticle body has a small shroud protruding around both photoresistors, reducing the angle of acceptance of the controlling light source and slightly modifying the geometry of the smarticle case. The shrouded smarticles were controlled with a static bar of point light sources located at the center of the edge of the test plate. The reduced angle of acceptance and internal ensemble geometry lead to the nearest smarticle entering the inactive state. The shrouded implementation gives a similar internal material behavior to the exposed smarticle system while removing the need for a manually controlled light source. All supersmarticles have homogeneous populations of smarticles: all smarticles within the ensemble were either exposed or shrouded.

For both forms of supersmarticles, two types of experiments were performed: one in which no light source existed, such that all smarticles remained active, and the second in which a controlling light source was introduced, such that both active and inactive smarticles could emerge in response to signals from the photosensors. All experiments were performed in a dark room, so that smarticles only entered the inactive state when subjected to the controlling light source input. Experimental trials were initiated with the supersmarticle system near the center of a flat, leveled 0.2 x 0.2 $m^2$ test plate, which ended when the supersmarticle translated to an edge of the test plane. When internal supersmarticle configurations exhibited slow displacement rates, trials terminated at 10 minutes. Experiments were repeated with the light source located at four different locations about the test plane to account for possible variances in results due to potential irregularities in the test surface.

Trajectories of the supersmarticle center of geometry were recorded using OptiTrack infrared video recording technology, and the data were exported and analyzed in MATLAB using a MSD analysis package \cite{Tarantino2014}.

\section{Results and Discussion}

\begin{figure}[htbp]
\centerline{\includegraphics[width=.5\textwidth]{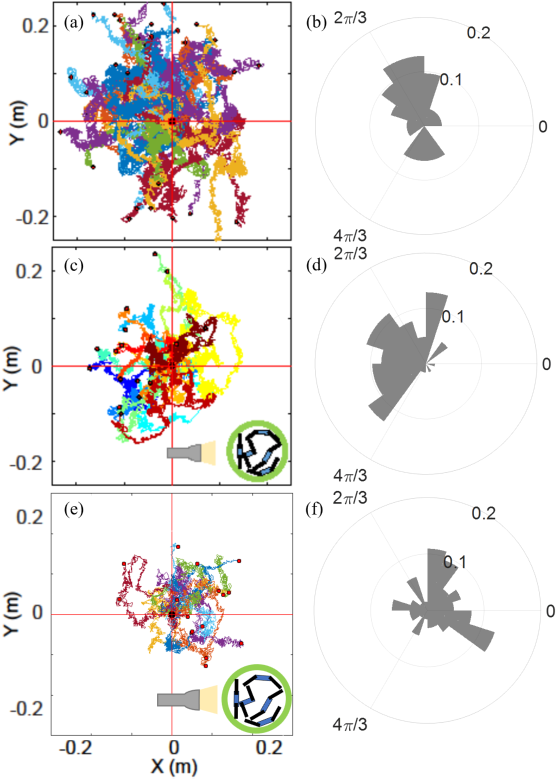}}
\caption{(a), (c), and (e) are trajectories of the supersmarticle’s center of geometry for a non-biased motion and light biased motion with exposed and shrouded smarticles, respectively. Each color represents a separate trial. (d) and (f) contain all exposed and shrouded smarticle light biased directions data. Trials where the light was not originating from the +X direction were rotated as though they were to allow us to compare more trials. Illumination direction is shown via the location of the flashlight with respect to the supersmarticle image. (b), (d), and (f) are polar histograms obtained for final supersmarticle displacement.}
\label{fig:tracks}
\end{figure}

Supersmarticle displacement experiments were performed with varying relative light source locations. The +X direction bias is shown in Fig. \ref{fig:tracks}.(c, e). Diffusive behavior is observed in both the control (Fig. \ref{fig:tracks}.(a)) and directed experiments (Fig. \ref{fig:tracks}.(c,e)), but the presence of inactive smarticles near the light source introduces a biased drift with respect to the direction of the light. Polar histograms were produced from the net displacement of each supersmarticle trial, with 0 indicating supersmarticle displacement away from the light source and $\pi$ towards the light source. The exposed smarticle system consistently drifted in the direction of the light source, with 49/62 trials moving towards the light across all trials. The biased diffusion for the shrouded smarticle cases, contrary to the exposed smarticle ensembles, resulted in 21/64 trials moving towards the light.

\begin{figure}[htbp]
\centerline{\includegraphics[width=.35\textwidth]{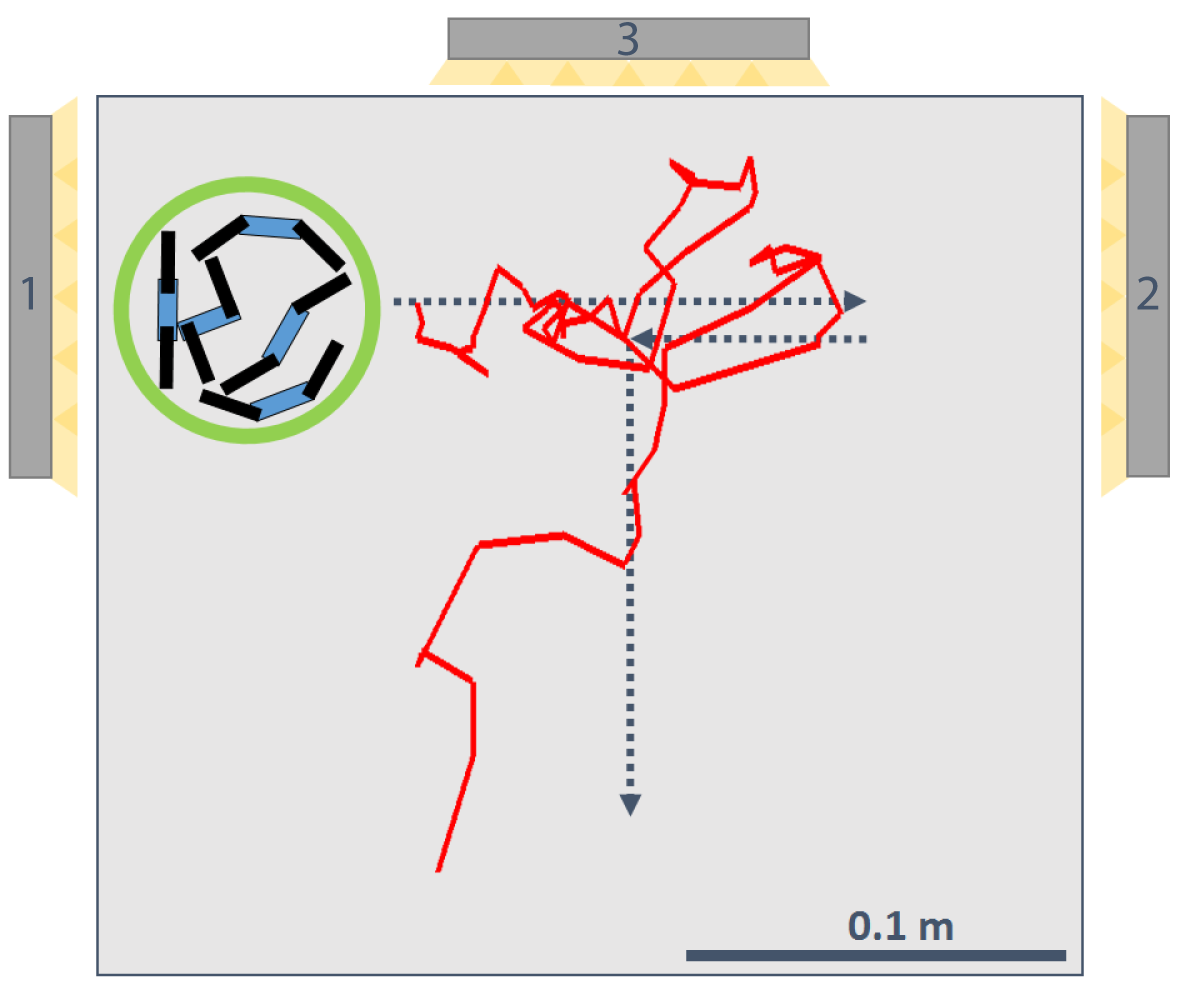}}
\caption{Dashed arrows outline the desired \textbf{T} trace, with the red line denoting the track of the supersmarticle as the ensemble is repelled by elements of the lighting configuration. Only one light source is on at any given time. The trial begins with Light 1 on, which repels the supersmarticle to the right. When the supersmarticle is near Light 2, Light 2 is activated to reverse the direction of diffusion back to the left side of the testbed. When the supersmarticle is near the midpoint of the plane, Light 3 is activated to repel the ensemble down in a path perpendicular to the two previous biased diffusion paths.}
\label{fig:t}
\end{figure}

For both smarticle forms, the phototaxing phenomenon was consistent, with exposed smarticle ensembles exhibiting positive phototaxing diffusion while the shrouded smarticles exhibited negative phototaxing. The phototaxing behavior was regular enough to be exploited to draw shapes within the plane of diffusion. The \textbf{T}, seen in Fig. \ref{fig:t} was created using the shrouded cases by changing the location of the repelling light source as the supersmarticle moved throughout the test plane.


\section{Conclusion}

This study presented the use of immobile agents in the formulation of a robotic material which exhibits locomotion on the collective scale. The direction of this locomotion is random when there are no biases within the robotic material, though internal asymmetries in response to external stimuli allow the material to exhibit a biased, diffusive locomotion. The change in the body geometry (i.e., photosensor shroud) had an effect on the direction of the biased diffusive behavior, which were found to be consistent for both the exposed and shrouded supersmarticle ensembles, suggesting that there is a physics governed by the interaction geometry that needs to be fully explored.

Future work will investigate the dynamics which the supersmarticle utilizes to move. We hypothesize that the unbiased direction of locomotion is random because there is no mechanism to consistently precipitate a particular family of interaction modes from trial-to-trial. There is no consistent geometric asymmetry in the unbiased ensembles, such that modes of interactions which propagate in a given direction can emerge in any orientation. For supersmarticle systems in which a consistent bias is produced by exposure to a light source, the ability for modes of smarticle interactions to emerge in a consistent direction is facilitated. The physics of the biased locomotion we have presented will be more thoroughly explored in \cite{Savoie2017}, accompanied by the presentation of several other robotic materials composed of smarticle ensembles.


\printbibliography[title={References}]

\end{document}